\documentclass[letterpaper, 10 pt, conference]{ieeeconf}  
\IEEEoverridecommandlockouts                              
\usepackage{graphics} 
\usepackage{epsfig} 
\usepackage{mathptmx} 
\usepackage{times} 
\usepackage{amsmath} 
\usepackage{amssymb}  
\usepackage{siunitx}

\title{\LARGE \bf
Ori-Sense: origami capacitive sensing for soft robotic applications}

\author{Hugo de Souza Oliveira*, Xin Li, Mohsen Jafarpour, and Edoardo Milana*%
\thanks{All authors are with livMatS @ FIT and the Department of Microsystems Engineering (IMTEK), University of Freiburg, 79110 Freiburg, Germany.}%
\thanks{Corresponding authors: hugo.oliveira@livmats.uni-freiburg.de, milana@imtek.de}%
}

\begin{document}

\maketitle
\thispagestyle{empty}
\pagestyle{empty}

\begin{abstract}

This work introduces Ori-Sense, a compliant capacitive sensor inspired by the inverted Kresling origami pattern. The device translates torsional deformation into measurable capacitance changes, enabling proprioceptive feedback for soft robotic systems. Using dissolvable-core molding, we fabricated a monolithic silicone structure with embedded conductive TPU electrodes, forming an integrated soft capacitor. Mechanical characterization revealed low stiffness and minimal impedance, with torque values below \SI{0.01}{\newton \milli \metre} for axial displacements between \SI{-15}{\milli\metre} and \SI{15}{\milli\metre}, and up to \SI{0.03}{\newton \milli\metre} at \SI{30}{\degree} twist under compression. Finite-element simulations confirmed localized stresses along fold lines and validated the measured torque–rotation response. Electrical tests showed consistent capacitance modulation up to 30 \%, directly correlated with the twist angle, and maximal sensitivity of $S_\theta \approx \SI{0.0067}{\pico \farad \per \degree}$ at \SI{5}{\milli\metre} of axial deformation. The inverted Kresling geometry provides decoupled deformation modes, high compliance, and robust electrode integration, making Ori-Sense a promising platform for embedded proprioception in soft actuators, wearable systems, and multifunctional robotic materials.

\end{abstract}


\section{Introduction}

In the last decade, soft robotics has emerged as a complementary pathway to conventional robotics for a variety of tasks, ranging from medical surgery to the manipulation of delicate objects~\cite{ Navas_2023}. Using compliant materials instead of rigid components, soft robots are uniquely suited for unstructured environments and for working alongside humans, as their inherent compliance can mimic the mechanical properties of biological tissues, translating to safer interactions~\cite{van_2025}. Several design approaches have been explored in the fabrication of soft robots, often drawing inspiration from nature to create actuators and structures capable of complex, continuous deformations that are unattainable for their rigid counterparts.

To transform these compliant machines into autonomous robots, it is essential to integrate them with equally compliant electronic and sensing capabilities. The field of soft electronics has risen to this challenge, developing sensors that can stretch, twist, and bend along with the robot's body~\cite{Oliveira_2025a}. The goal is to create a seamless electronic "skin" or an integrated nervous system that endows soft robots with proprioception (self-sensing of internal state) and exteroception (perception of the environment), without compromising their fundamental softness~\cite{Wang_2025}.

For soft robots to operate autonomously, such sensory feedback mechanisms are essential. Most soft machines currently operate in an open-loop mode, with their behaviors programmed by design but lacking the ability to adapt to unforeseen interactions~\cite{milana2022morphological}. Integrating sensors is essential for closing the control loop, but traditional soft sensors often face challenges with hysteresis, complex fabrication, and difficulty in decoupling different modes of deformation, such as bending versus twisting deformation. To address these limitations, mechanical metamaterials—materials whose unique properties derive from their engineered structure, rather than from the properties of the material—have been realized~\cite{Oliveira_2025a}. By designing specific geometries, it is possible to create materials with programmable responses, offering a powerful approach to enhance sensor functionality.

\begin{figure}[t]
    \centering
    \includegraphics[width=0.45\textwidth]{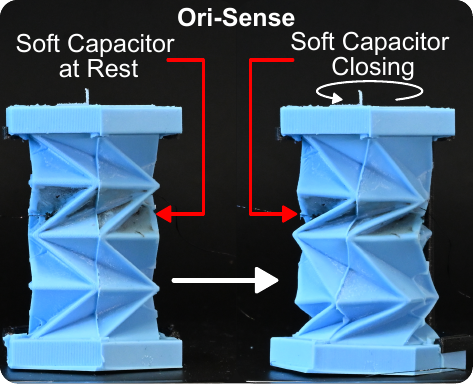}
    \caption{Concept of the Ori-Sense soft origami capacitor. The inverted Kresling structure translates torsional deformation into a measurable capacitance change, enabling proprioceptive sensing in soft robotic systems.}
    \label{fig-fancy}
\end{figure}

Measuring complex deformation modes such as torsion remains a major challenge in soft robotics. Conventional soft sensors often mix multiple stimuli, making it difficult to distinguish, for example, twisting from stretching or bending from compression. Architected materials inspired by kirigami and origami overcome this limitation by using fold and cut patterns that can be tailored so that specific mechanical inputs, like a twist, produce well-defined local deformations. In these systems, the geometry itself acts as the transduction element, enabling precise and selective sensing of complex mechanical states.

 \begin{figure*}[htbp]
     \centering
     \includegraphics[width=0.9\textwidth]{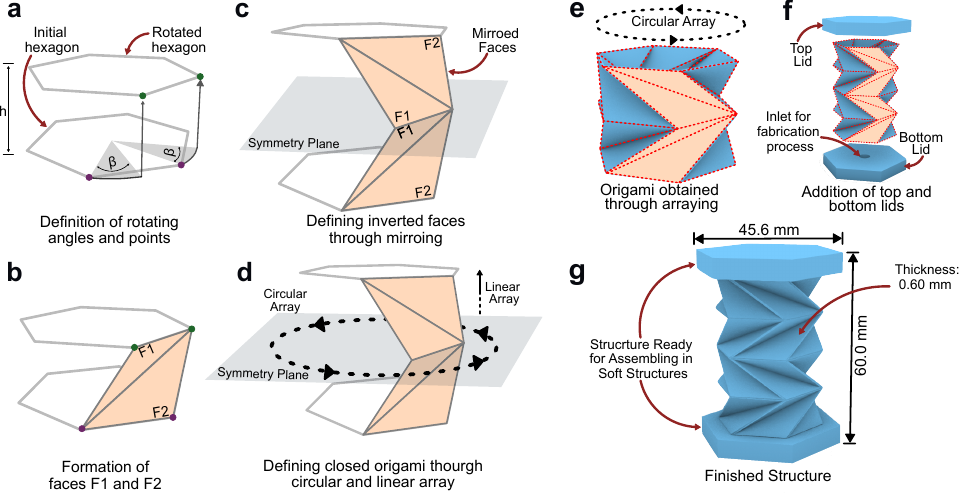}
     \caption{Design of the origami structure: (a) Initial hexagon and point rotation by $\beta$. (b) Formation of faces $F_1$ and $F_2$. (c) Mirroring through a symmetry plane. (d) Circular array construction. (e) Linear array extension. (f) Addition of top and bottom lids. (g) Final assembled structure.}
     \label{fig-design}
 \end{figure*}

In this work, we introduce Ori-Sense, an origami-inspired capacitive sensing device that provides proprioceptive capabilities for soft robotic actuators. Differently from existing approaches that uses normal kresling origami deformation~\cite{Hanson_2024, Jiao_2023}, we explore an inverted kresling design, leveraging the principles of mechanical metamaterials by employing a specific origami fold pattern that translates rotation into localized, measurable changes in capacitance. This structured design allows for the decoupling of complex deformations, providing a sensitive measurement of a soft actuator's configuration. This work represents a crucial step toward creating fully compliant, self-aware robotic systems capable of feedback control.

\section{Design, Fabrication \& Methods}

\subsection{Origami design}

\begin{figure*}[htpb]
    \centering
    \includegraphics[width=0.9\textwidth]{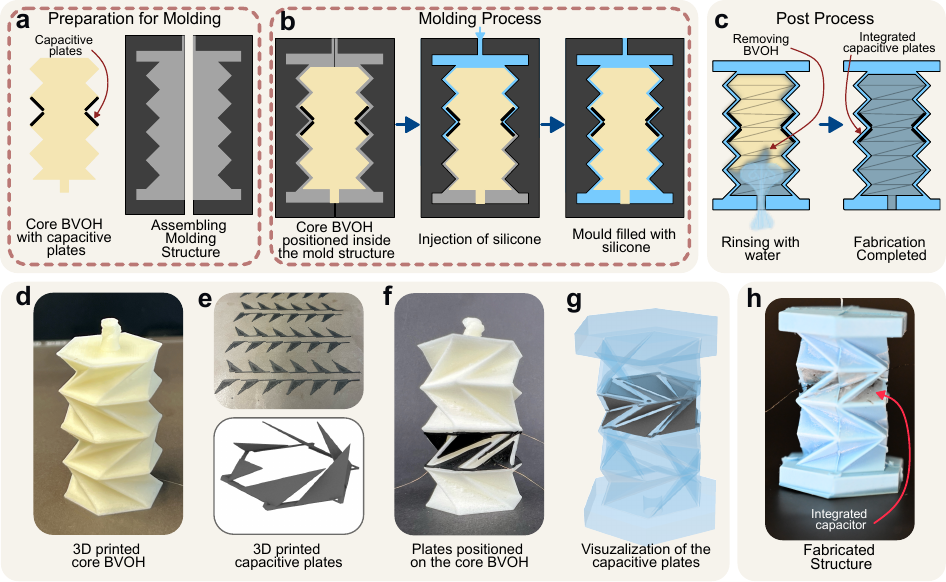}
    \caption{Fabrication of the soft origami capacitive sensor: (a) Printing of BVOH core, addition of the capacitive plates and molding prepararton. (b) Mold assembly and silicone injection. (c) Core dissolution after curing. (d) 3D-printed core. (e) Conductive TPU plates. (f) Core with plates before molding. (g) Capacitive plates arrangement in silicone. (h) Final soft structure with integrated electrodes.}
    \label{fig-fabrication}
\end{figure*}

Fig.~\ref{fig-design} depicts the development process of the origami capacitive sensor derived from a Kresling pattern. The design procedure starts by defining a regular hexagon with two points located on one of its sides (Fig.~\ref{fig-design}a). These points are then rotated about the hexagon center by an angle $\beta$, as shown by the purple marker in Fig.~\ref{fig-design}a. Subsequently, they are lifted by a height $h$, corresponding to the Kresling pattern height, to form a rotated hexagonal layer indicated by the green points in Fig.~\ref{fig-design}a. Connecting the green and purple points defines the first two triangular panels of the Kresling structure, $F_1$ and $F_2$, as illustrated in Fig.~\ref{fig-design}b. 

A symmetry plane parallel to the base hexagon and offset by the same height $h$ is used to mirror these faces (Fig.~\ref{fig-design}c), yielding the inverted Kresling configuration. This inversion compensates for the rotational motion induced by the folding of the Kresling pattern during axial deformation and is essential in the deformation effect demonstrated in Fig.~\ref{fig-fancy} and in the supplemental Video. The complete origami structure is obtained through a combination of circular and linear array operations Fig.~\ref{fig-design}d, leading to the final pattern shown in Fig.~\ref{fig-design}e. Top and bottom lids are added to the origami structure to serve as holders transmitting axial and rotational deformations to the origami structure, as seen in Fig.~\ref{fig-design}f. The bottom lid has an entrance to allow air to flow in/out when the structure is deformed and to allow for water flow during the fabrication process. The resulting structure (Fig.~\ref{fig-design}g) has overall dimensions of \SI{45.6}{\milli\metre} $\times$ \SI{45.6}{\milli\metre} $\times$ \SI{60.0}{\milli\metre}$,$ with an origami wall thickness of \SI{0.60}{\milli\metre}, a folding angle $\beta$ of \SI{75}{\degree}, and an offset $h$ of \SI{6.0}{\milli \meter}.

\subsection{Fabrication process} 

Figure~\ref{fig-fabrication} depicts the fabrication process of the soft capacitive origami-sensor. Figures~\ref{fig-fabrication}a–c show a schematic representation of the dissolvable-core molding process. The fabrication begins with the 3D printing of a core made from a water-soluble Butenediol Vinyl Alcohol Co-polymer (BVOH) using a multi-material fused deposition modeling (FDM) printer (Prusa XL) using standard printing speeds and accelerations (Figure~\ref{fig-fabrication}a). Conductive TPU plates are printed separately to serve as capacitive electrodes and are subsequently attached to the slanted faces of the BVOH core using double-sided adhesive tape. Then, thin copper wires are glued in the capacitive plates to serve as electrical contacts. The prepared core is then placed inside a four-piece assembling mold designed to tightly enclose it during silicone casting (Figure~\ref{fig-fabrication}b). Liquid silicone is injected into the assembled mold, encapsulating the BVOH core fully and filling the surrounding cavity. After curing, the structure is removed from the mold, completely immersed in water and sonicated for \SI{4}{\hour} at \SI{50}{\celsius} to dissolve the BVOH, resulting in a hollow silicone body with conductive TPU plates embedded on its internal surfaces (Figure~\ref{fig-fabrication}c).

Figures~\ref{fig-fabrication}d–g present details of the main steps of the fabrication process. Figure~\ref{fig-fabrication}d shows the 3D-printed BVOH core, while Figure~\ref{fig-fabrication}e illustrates the conductive TPU plates in their printed form and their conformed geometry once attached around the core. Figure~\ref{fig-fabrication}f displays the assembled core with the conductive plates positioned prior to molding, and Figure~\ref{fig-fabrication}g provides a schematic view of the final electrode arrangement within the silicone structure. The completed sensor, shown in Figure~\ref{fig-fabrication}h, features integrated capacitive plates enclosed in a compliant silicone shell. The device operates as a soft capacitive sensor that leverages the characteristic torsional deformation of the Kresling origami (Figure~\ref{fig-fancy}, where opposite faces move closer together and induce a measurable change in capacitance, making it suitable for deformation sensing in soft robotic systems.

\subsection{Mechanical characterization}

The structure was characterized under combined torque and axial deformation using a uniaxial testing machine (ZwickRoell Z010) capable of applying both torque and tensile or compressive forces simultaneously. The specimen was first compressed by \SI{15}{\milli\meter} and then subjected to ten rotational cycles between \SI{0}{\degree} and \SI{30}{\degree}. The axial compression was subsequently released in increments of \SI{5}{\milli\meter} until a tensile elongation of \SI{15}{\milli\meter} was reached. For each axial deformation level, ten rotation cycles were again performed between \SI{0}{\degree} and \SI{30}{\degree}. During all tests, both the axial force and the applied torque were continuously logged for each rotation cycle.

\begin{figure*}[htbp]
     \centering
     \includegraphics[width=0.9\textwidth]{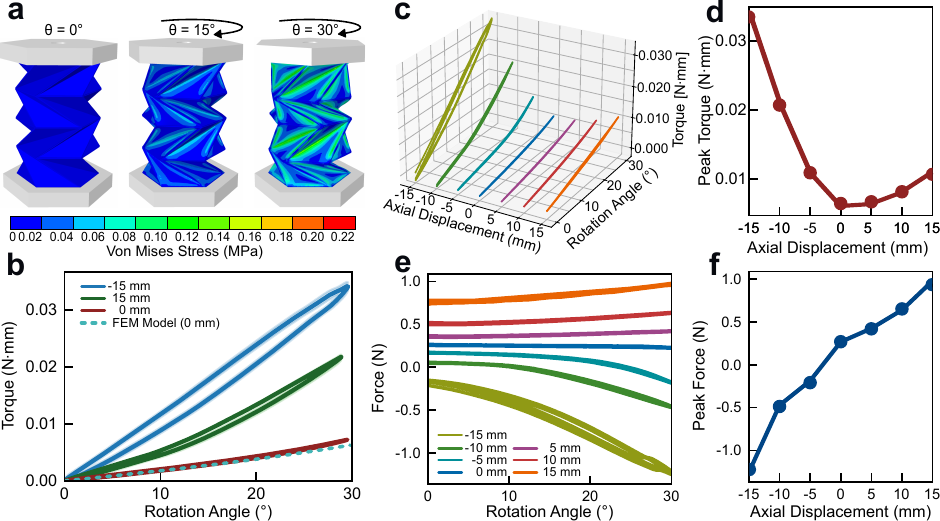}
     \caption{Mechanical response of the inverted Kresling origami sensor:
(a) Von~Mises stress distribution at \SI{30}{\degree} of rotation. (b) Comparison between experimental and FEM torque–rotation curves at zero axial strain, together with extreme cases. (c) Torque–rotation curves for varying axial displacements. (d) Peak torques for the structure. 
(e) Axial force versus rotation angle for different axial offsets. 
(f) Peak axial force and torque, indicating symmetric trends between tensile and compressive deformation.}
     \label{fig-mech}
 \end{figure*}

\subsection{Electrical Characterization}

The signals generated by the flexible capacitor was acquired using a Texas Instruments FDC2214EVM board, which integrates the \textit{FDC2214} chip—a 28-bit capacitance-to-digital converter~\cite{Oliveira_2025b}. The capacitor was connected to one of the four input channels of the board. Instead of directly measuring capacitance, the FDC2214EVM operates an LC (inductor–capacitor) tank circuit whose resonant frequency varies with the spacing between the capacitor plates. Although the conductive TPU electrodes exhibit higher resistivity than conventional metal electrodes, their conductivity remains sufficient at the device’s operating frequency to keep the alternating electric field required for resonance. 

The FDC2214 excites the LC circuit, detects the resulting frequency shift, and converts it into a digital signal. Variations in the plate separation alter the effective capacitance, thereby changing the resonance frequency, which is translated by the board into a proportional digital count. Due to the chip’s high sensitivity, the raw output contains a significant static offset, masking small fluctuations. To compensate for this, a custom acquisition script performs continuous dynamic calibration by tracking the real-time minimum and maximum sensor values. The instantaneous signal is offset-corrected by subtracting the current minimum and subsequently normalized to a 0–100 range, enabling clear visualization of small variations while automatically compensating for slow environmental drifts.

\subsection{Finite-element simulations} 

The mechanical response of the origami structure was analyzed using finite-element simulations conducted in Abaqus/Standard, v. 2025 (Dassault Systèmes Simulia Corp., Providence, RI, USA). The CAD geometry of the origami unit was imported and meshed with C3D10HS quadratic tetrahedral elements, using a global element size of \SI{0.5}{\milli\metre} and curvature control enabled to refine the mesh near fold intersections. A mesh sensitivity analysis was performed to ensure that the numerical results were independent of the element size, confirming the convergence of both the deformation pattern and the torque–rotation response.

The Smooth-Sil 950 silicone material was modeled as an isotropic Neo-Hookean hyperelastic solid with parameters C10 = \num{0.34} and D1 = \num{0}, and a density of \SI{1240}{\kilogram\per\cubic\metre}, based on values reported in previous studies \cite{xavier2021finite, gariya2022stress}. Geometrical non-linearity was included in the simulations, and self-contact was defined between all surfaces using a surface-to-surface discretization method to prevent interpenetration during twisting.

A dynamic implicit analysis step was employed to simulate quasi-static torsion of the origami structure. The bottom surface of the model was fixed in all translational and rotational directions, while an angular displacement of \SI{30}{\degree} was applied to the top surface about the central axis, as shown in Figure~\ref{fig-elec}b. The reaction torque and the rotational displacements were recorded to obtain the torque–rotation relationship and to visualize the deformation process.
\section{Results \& Discussion}

The inverted Kresling configuration employed here (Fig.~\ref{fig-design}e) introduces a mirrored fold topology in which the triangular panels are mirrored across a symmetry plane, as shown in Fig.~\ref{fig-design}c. This inversion fundamentally alters the kinematics of the structure: as the origami twists, opposing faces counteract each other, leading to a reduction in both axial displacement and reaction force (Supplementary Video, Fig.~\ref{fig-fancy}). The resulting deformation is highly characteristic, as one pair of faces ($F_1$, $F_2$) moves toward each other while the adjacent pair separates, producing an alternation of compression and expansion along the axial direction. This coupled motion forms the geometric basis for the capacitance variation observed in the structure.

\begin{figure*}[htbp]
     \centering
     \includegraphics[width=0.9\textwidth]{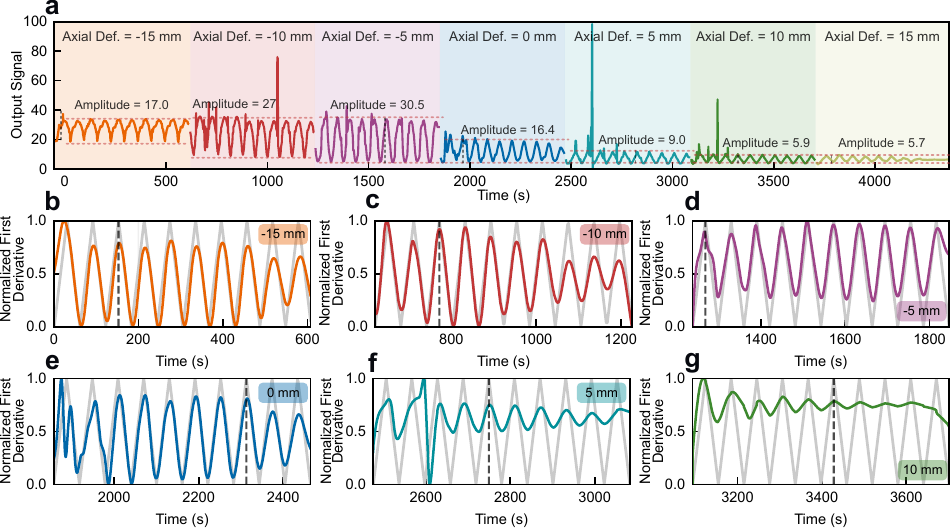}
     \caption{Electrical characterization of the soft origami capacitive sensor. 
(a) Normalized sensor output during repeated torsional cycles (\SI{0}{\degree}–\SI{30}{\degree}) for axial displacements of \SIlist{-15;-10;-5;0;5;10;15}{\milli\meter}. 
(b) Normalized signal for the case of \SI{-15}{\milli\meter} axial displacement. 
(c) Normalized signal for \SI{-10}{\milli\meter}. 
(d) Normalized signal for \SI{-5}{\milli\meter}. 
(e) Normalized signal for \SI{0}{\milli\meter}. 
(f) Normalized signal for \SI{10}{\milli\meter}. 
(g) Normalized signal for \SI{15}{\milli\meter}.}
     \label{fig-elec}
 \end{figure*}

A key challenge in soft capacitive sensing is maintaining adhesion between conductive layers and elastomeric substrates. In conventional soft capacitors, mechanical mismatch and distributed strain often promote delamination of thin conductive films. In contrast, the Kresling geometry localizes stress primarily along the fold lines~\cite{Hanson_2024}, as confirmed by the finite-element simulation (Fig.~\ref{fig-mech}a). The simulation reproduced the progressive torsional deformation observed experimentally. The structure exhibited a uniform rotational motion along its central axis, accompanied by localized folding near the midsection, consistent with Kresling-type kinematics. During twisting, opposite faces alternately approached or moved apart, reflecting the coupled axial and rotational deformation that defines this geometry. The von~Mises stress distribution revealed clear stress concentrations along the fold lines and hinge regions, whereas the flat surfaces bearing the capacitive electrodes experienced only minor stresses. This confirms that the structural design effectively isolates the sensing areas from significant mechanical load, minimizing detachment or damage.

The simulated torque–rotation curve closely matched the experimental measurements, following the same nonlinear trend with a smooth increase in torque up to approximately \SI{0.06}{\newton\milli\metre} at \SI{30}{\degree}, validating the accuracy of the finite-element model (Fig.~\ref{fig-mech}b). This agreement supports the mechanical interpretation of the origami behavior and provides a computational tool for exploring alternative geometries, materials, and loading conditions in future studies.

The fabrication approach further reinforces the mechanical stability of the sensing interfaces. During molding, uncured silicone infiltrates the micro-texture of the conductive TPU plates (Fig.~\ref{fig-fabrication}b,c), ensuring interlocking between the layers. This process yields a monolithic composite in which the electrodes are effectively embedded rather than laminated, significantly enhancing durability under torsional loading.

Figures~\ref{fig-mech}c–f summarize the mechanical response under combined axial and torsional deformation. For axial deformations (\SIrange{-5}{15}{\milli\metre}), the torque required to reach \SI{30}{\degree} remains below \SI{0.01}{\newton\meter}, indicating a mechanically compliant range that can be explored to integrate into into soft robotic and machines. At larger compressive displacements (\SI{-10}{\milli\metre} and \SI{-15}{\milli\metre}), the origami folds collapse and frictional contact between faces dominates, raising the torque to \SIrange{0.02}{0.03}{\newton\meter}. Under these locked conditions, further twisting becomes restricted. The axial-force measurements (Fig.~\ref{fig-mech}e) reveal nearly constant values for neutral and tensile states, with small offsets attributed to assembly tolerances. In contrast, under compression, the force increases sharply beyond \SIrange{10}{15}{\degree} rotation as frictional contact develops. The resulting force peaks (Fig.~\ref{fig-mech}f) exhibit symmetric trends between tensile and compressive cases. These results confirm that the structure exerts minimal load in its operational range—a. This is a desirable attribute when the sensor is mounted on compliant actuators, as it will minimally change its mechanical properties.

Figure~\ref{fig-elec}a presents the normalized capacitive response acquired with the FDC2214EVM and the custom acquisition script. Each curve corresponds to rotational cycles between \SI{0}{\degree} and \SI{30}{\degree} for a given axial displacement. Across all conditions, the capacitance variation follows the rotation, indicating consistent electromechanical coupling. The \SI{5}{\milli\metre} condition yields the largest modulation for the adopted geometry, suggesting an optimal balance between approach and separation of the plates during twisting. Under extreme compression (\SI{-15}{\milli\metre}), the structure becomes nearly locked and opposite faces are almost in contact, leading to a high baseline capacitance but a reduced dynamic range; as a result, the observed amplitude remains comparable to the undeformed case (\SI{0}{\milli\metre}). For positive axial deformations (\SIrange{10}{15}{\milli\metre}), the plate spacing increases and the signal modulation decreases. Overall, the capacitive response can be tuned via geometric pre-strain, adjusting sensitivity without changing material composition. For the case of \SI{0}{\milli\metre} axial deformation, the capacitance ranges from approximately \SI{0.1}{\pico\farad} at \SI{0}{\degree} to \SI{0.3}{\pico\farad} at \SI{30}{\degree}. From this capacitance variation, the average angular sensitivity can be estimated as $S_\theta \approx \SI{0.0067}{\pico \farad \per \degree}$. The occasional sharp peaks observed in Figure~\ref{fig-elec}a are attributed to environmental electrical disturbances. This value represents the effective electromechanical coupling of the inverted Kresling geometry under torsional deformation.

Figures~\ref{fig-elec}b–g show the time derivative of the normalized signal together with the measured rotation angle. The capacitance peaks coincide with the maximum twist (\SI{30}{\degree}), confirming a direct correlation between torsional deformation and electrical output. This correlation demonstrates that the sensor can accurately encode the rotational state, providing proprioceptive feedback for soft machines. Ongoing work aims to extend this concept toward dual-mode sensing by introducing additional electrode configurations capable of simultaneously capturing torsional and axial strain.

Overall, the inverted Kresling origami provides a geometrically decoupled, mechanically compliant, and electrically stable platform for soft capacitive sensing. Its localized strain fields, low mechanical impedance, and embedded-electrode architecture make it highly suitable for integration into soft machines and wearable systems, offering a new route for proprioceptive sensing in compliant robotics~\cite{Oliveira_2025b}.

\section{Conclusion and Outlook}

In this work, we presented the design, fabrication, and characterization of \textit{Ori-Sense}, a compliant capacitive sensor based on an inverted Kresling origami pattern. The proposed architecture effectively converts torsional deformation into measurable capacitance changes, enabling direct proprioceptive feedback of the rotational state. Mechanical testing and finite-element simulations confirmed that the inverted Kresling geometry localizes strain along the fold lines while maintaining extremely low stiffness across its operational range. Torque values remained below \SI{0.01}{\newton\milli \meter} for axial offsets between \SIrange{-15}{15}{\milli\meter}, and only reached \SI{0.03}{\newton \milli \metre} under compression at \SI{30}{\degree} of twist, demonstrating that the structure can deform with minimal resistance.

This low mechanical resistance is a central feature of the design. It ensures that the sensor does not interfere with the deformation or actuation of the host soft system, allowing the robot to retain its intrinsic compliance. In practical terms, this means that \textit{Ori-Sense} can be embedded within soft actuators or wearable structures without altering their mechanical response, while still providing accurate information about their configuration. This combination of sensitivity and softness is difficult to achieve with traditional stretchable or laminated sensors, which often stiffen the system or suffer from delamination and hysteresis.

The inverted Kresling topology therefore represents a promising approach toward truly integrated soft sensing—one that exploits geometry rather than material stiffness to achieve signal transduction. Beyond torsion detection, the structure’s geometrically coupled axial–rotational behavior provides a basis for dual-mode sensing, where both twist and extension can be captured through additional capacitive channels. Because of its low actuation torque and intrinsic mechanical decoupling, \textit{Ori-Sense} opens new possibilities for closed-loop control in soft machines that are composed entirely of compliant materials. 

Future work will focus on developing a complete electromechanical model fully correlating capacitance and axial strain, as well as embedding the sensor directly into pneumatic and origami-based actuators to demonstrate autonomous feedback and motion control. 


\section*{Acknowledgment}
This work was funded by the Deutsche Forschungsgemeinschaft (DFG, German Research Foundation) under Germany’s Excellence Strategy-EXC-2193/1-390951807.

\bibliographystyle{ieeetr}
\bibliography{references}
\end{document}